\pdfoutput=1

\documentclass[11pt]{article}

\usepackage{EMNLP2022}

\usepackage{color,soul}
\usepackage{times}
\usepackage{latexsym}
\usepackage{url}
\usepackage{amsmath}
\usepackage{amssymb}
\usepackage{multirow}
\usepackage{float}
\DeclareMathOperator*{\argmax}{arg\,max}

\usepackage{enumitem}

\usepackage{graphicx}
\graphicspath{ {./images/} }

\usepackage[T1]{fontenc}

\usepackage[utf8]{inputenc}

\usepackage{microtype}

\title{Topical Segmentation of Spoken Narratives:\\ A Test Case on Holocaust Survivor Testimonies}

\author{Eitan Wagner$^\dagger$\quad Renana Keydar$^\ddagger$\quad Amit Pinchevski$^\diamondsuit$ \quad Omri Abend$^\dagger$ \\
         $^\dagger$ Department of Computer Science \quad
         $^\ddagger$ Faculty of Law and Digital Humanities\\
         $^\diamondsuit$ Department of Communication and Journalism\\
         Hebrew University of Jerusalem\\ \texttt{\{first\_name\}.\{last\_name\}@mail.huji.ac.il}}

\begin{document}
\maketitle

\begin{abstract}

The task of topical segmentation is well studied, but previous work has mostly addressed it in the context of structured, well-defined segments, such as segmentation into paragraphs, chapters, or segmenting text that originated from multiple sources. We tackle the task of segmenting running (spoken) narratives, which poses hitherto unaddressed challenges. 
As a test case, we address Holocaust survivor testimonies, given in English. Other than the importance of studying these testimonies for Holocaust research, we argue that they provide an interesting test case for topical segmentation, due to their unstructured surface level, relative abundance (tens of thousands of such testimonies were collected), and the relatively confined domain that they cover.
We hypothesize that boundary points between segments correspond to low mutual information between the sentences proceeding and following the boundary. Based on this hypothesis, we explore a range of algorithmic approaches to the task, building on previous work on segmentation that uses generative Bayesian modeling and state-of-the-art neural machinery. 
Compared to manually annotated references, we find that the developed approaches show considerable improvements over previous work.\footnote{Code is provided at  \url{https://github.com/eitanwagner/holocaust-segmentation}}

\end{abstract}

\section{Introduction}



\begin{figure}[t]
\centering
\includegraphics[scale=0.5]{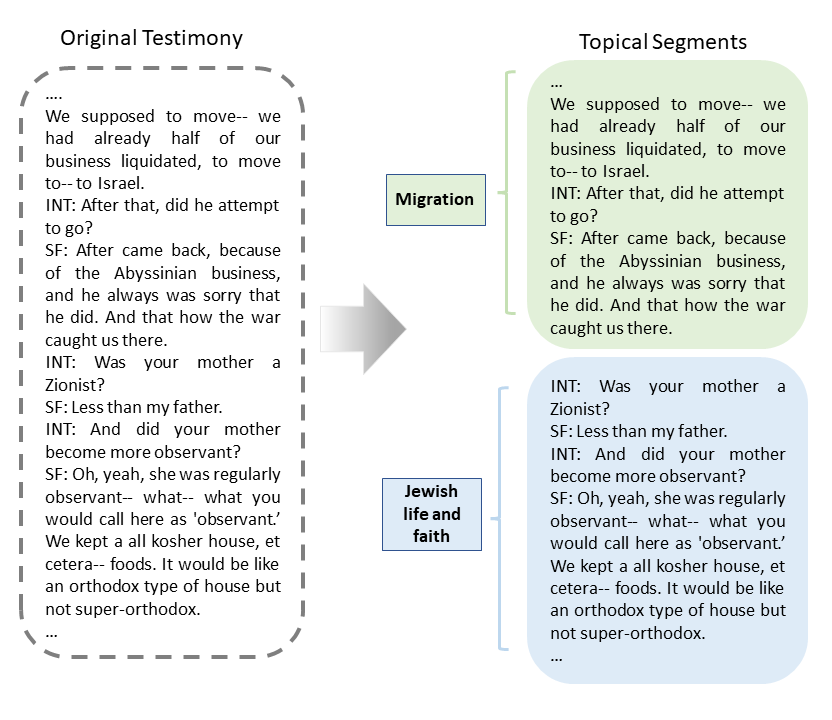}
\caption{Topical segmentation in Holocaust testimonies.}
\end{figure}

Proper representation of narratives in long texts remains an open problem in NLP \citep{pipernarrative, castricato-etal-2021-towards, mikhalkova-etal-2020-modelling}. High-quality representations for long texts seem crucial to the development of document-level text understanding technology, which is currently unsatisfactory \citep{shaham2022scrolls}.
A common modern approach for modeling narratives is as a sequence of neural states \citep{wilmot-keller-2020-modelling, DBLP:journals/corr/abs-2109-06807, rashkin-etal-2020-plotmachines}. However, a drawback of this approach is the lack of interpretability, which is crucial in some contexts.

A different approach represents and visualizes a narrative as a sequence of interpretable topics \cite{antoniak2019narrative}.
Inspired by this approach, we seek to model the narrative of a text using topic segmentation, dividing long texts into topically coherent segments and labeling them, thus creating a global topical structure in the form of a chain of topics. 
Topic segmentation can be useful for the indexing of a large number of testimonies (tens of thousands of testimonies have been collected thus far) and as an intermediate or auxiliary step in tasks such as summarization \citep{wu2021recursively} and event detection \citep{wang-etal-2021-learning-constraints}.

Unlike recent supervised segmentation models that focus on structured written text, such as Wikipedia sections \citep{arnold-etal-2019-sector, lukasik-etal-2020-text} or book chapters \citep{pethe-etal-2020-chapter}, we address the hitherto mostly unaddressed task of segmenting and labeling unstructured (transcribed) spoken language. 
For these texts, we don't have large datasets of divided text. Moreover, there may not be any obvious boundaries that can be derived based on local properties. This makes the task more challenging and hampers the possibility of taking a completely supervised approach. 

We propose an unsupervised alternative for segmentation, based on two assumptions: (1) segment boundaries correspond to places with low mutual information between sentences over the boundary; (2) neural language models can serve as reliable sentence probability estimators. Based on these assumptions, we propose a simple approach to segmentation and offer extensions involving dynamic programming. The proposed models give a substantial margin over the existing methods in terms of segmentation performance. 
In order to adapt the model to jointly segment and classify, we incorporate into the model a supervised topic classifier, trained over manually indexed one-minute testimony segments, provided by the USC Shoah Foundation (SF).\footnote{\url{https://sfi.usc.edu/}} Inspired by \citet{misra2011text}, we also incorporate the topical coherence based on the topic classifier into the segmentation model.

Our contributions are the following: (1) we present the task of topical segmentation for running, unedited text; (2) we propose novel algorithmic methods for tackling the task without any manual segmentation supervision, building on recent advances in language modeling; (3) comparing to previous work, we find substantial improvements over existing methods; (4) we compile a test set for evaluation in the case of Holocaust testimonies; (5) we develop domain-specific topical classifiers to extract lists of topics for long texts.

Typically, narrative research faces a tradeoff between the number of narrative texts, which is important for computational methods, and the specificity of the narrative context, which is essential for qualitative narrative research \citep{sultananarrative}. Holocaust testimonies provide a unique case of a large corpus with a specific context.
Our work also communicates with Holocaust research, seeking methods to better access testimonies as the survivor generation is slowly passing away \citep{artstein-etal-2016-new}. We expect our methods to promote schema-based analysis and browsing of testimonies, enabling better access and understanding.

\section{Previous work}

\paragraph{Text Segmentation.}
Considerable previous work addressed the task of text segmentation, using both supervised and unsupervised approaches. Proposed methods for unsupervised text segmentation can be divided into linear segmentation algorithms and dynamic graph-based segmentation algorithms.
    
Linear segmentation, i.e., segmentation that is performed on the fly, dates back to the TextTiling algorithm \citep{hearst1997text}, which detects boundaries using window-based vocabulary changes.  Recently, \citet{he2020improvement} proposed an improvement to the algorithm, which, unlike TextTiling, uses the vocabulary of the entire dataset and not only of the currently considered segment. TopicTiling \citep{riedl-biemann-2012-topictiling} uses a similar approach, using LDA-based topical coherence instead of vocabulary only. This method produces topics as well as segments.
Another linear model, BATS \citep{wu2020bats}, uses combined spectral and agglomerative clustering for topics and segments.
    
In contrast to the linear approach, several models follow a Bayesian sequence modeling approach, using dynamic programming for inference.
This approach allows making a global prediction of the segmentation, at the expense of higher complexity. Implementation details vary, and include using pretrained LDA models \citep{misra2011text}, online topic estimation \citep{eisenstein-barzilay-2008-bayesian,mota-etal-2019-beamseg}, shared topics \citep{jeong2010multi}, ordering-based topics \citep{du2015topic}, and context-aware LDA \citep{li2020context}.
    
Following recent advances in neural models, these models have been used for the task of supervised text segmentation. \citet{pethe-etal-2020-chapter} introduced ChapterCaptor which relies on two methods. The first method performs chapter break prediction based on Next Sentence Prediction (NSP) scores. The second method uses dynamic programming to regularize the segment lengths towards the average. The models use supervision for finetuning the model for boundary scores, but can also be used in a completely unsupervised fashion. They experiment with segmenting books into chapters, which offers natural incidental supervision.

Another approach performs the segmentation task in a completely supervised manner, similar to supervised labeled span extraction tasks.
At first, the models were LSTM-based \citep{koshorek-etal-2018-text, arnold-etal-2019-sector}, and later on, Transformer based  \citep{somasundaran2020two, lukasik-etal-2020-text}. Unlike finetuning, this approach requires a large amount of segmented data.

All of these works were designed and evaluated with structured written text, such as book chapters, Wikipedia pages, or artificially stitched segments, where supervised data is abundant. In this work, we address the segmentation of texts of which we have little supervised data regarding segment boundaries. We, therefore, adopt elements from the unsupervised approaches combined with supervised components and design a model for a novel segmentation task of unstructured spoken narratives.

\paragraph{Narrative analysis.}
Much work has been done in the direction of probabilistic schema inference, focusing on either event schemas \citep{chambers-jurafsky-2009-unsupervised, chambers-2013-event, li-etal-2020-connecting} or persona schemas \citep{bamman-etal-2013-learning, bamman-etal-2014-bayesian}.

Recently, neural models were utilized for story modeling. \citet{wilmot-keller-2020-modelling} presented a neural GPT2-based model for suspense in short stories. This work follows an information-based framework, modeling the reader's suspense by different types of predictability.
Due to their strong performance in text generation, neural models are commonly used for story generation, with numerous structural variations \citep{zhai-etal-2019-hybrid, rashkin-etal-2020-plotmachines, 10.1145/3453156}.
    
Narrative analysis can help in conveying the essence of stories, without all the details. This can aid the meta-analysis of stories. \citet{min2019modeling} visualized plot progressions in stories in various ways, including the progression of character relations. \citet{antoniak2019narrative} analyzed birth stories, using simplistic, uniform segmentation with topic modeling to visualize the frequent topic paths.

\section{Methods} \label{sec:methods}

We have a document $X$ consisting of $n$ sentences $x_{1}\cdots x_{n}$, which we consider as atomic units. Our task is to find $k-1$ boundary points, defining $k$ segments, and $k$ topics, where every consecutive pair of topics is different.



\vspace{-.03cm}
\subsection{Design Principles of Used Methods}

Designing a model for topical segmentation involves multiple, possibly independent, considerations which we present here.

\vspace{-.03cm}
\paragraph{Local Potential-Boundary Scores.}

A simple approach to text segmentation involves giving independent local scores to each possible boundary. Given these scores and the desired number of segments, we can then select the best boundaries. 

Recent work in this direction uses the Next Sentence Prediction (NSP) scores \cite{pethe-etal-2020-chapter}. 
Given two sentences $x_1,x_2$, their NSP score is defined as the predicted probability that the second sentence actually came after the first and not from somewhere else. The prediction is usually carried out using a pretrained model with a self-supervised training protocol and is typically further finetuned for a specific task.


We argue that the pretrained NSP scores do not capture the probability of two given sequential sentences being in the same segment, since even if the second sentence is in a new segment, it still is the next sentence. Therefore, we expect this approach to perform poorly in settings for which there are not enough segmented texts for finetuning.

 Instead, we propose to use Point-wise Mutual Information (PMI) for the local boundary scores. Given a language model (LM), we hypothesize that the mutual information between two adjacent sentences can predict how likely the two sentences are to be in the same segment. These scores need additional supervision beyond the LM pretraining. Given these scores, the extraction of a segmentation for a given text is equivalent to maximizing the LM likelihood of text, under the assumptions that each sentence depends on one previous sentence, and that each segment depends on no previous sentences (for proof see Appendix \ref{appendix: A}).

\paragraph{Non-local Scores.}
Full segmentation of text involves the selection of multiple boundaries, and these selections might not be independent. Even a single segment directly involves two boundaries. Therefore, we might want to use scores that take into account properties that involve more than one boundary. Given scores for all possible segments, we can optimize for the maximal total score over all possible segmentations.

A simple property that was used in previous work is the segment length \citep{pethe-etal-2020-chapter}, with a higher score given to segments whose length is closer to the expected length. These scores can be helpful if we assume that segments' length tends to be close to uniform. These scores can also be used in a conditional manner, in case we have estimates for the segment lengths of different topics or in different locations of the whole text. Segment length scores require the consideration of at least two corresponding boundaries for each score.

Another property that was used in previous work is topic scores \cite{misra2011text}. Given some Topic Model (TM), we can use the generation log-likelihood of a segment as its score. 
Alternatively, with supervised data for multi-label classification, we can use the classification log probabilities. With these scores, we can optimize for the maximal sum. 

Since we assume adjacent segments to have different topics, these scores must consider at least 3 boundaries creating two adjacent segments.


\paragraph{Pipeline, Joint Inference, Independent inference.}
The task of topical segmentation involves the extraction of both a segmentation and a corresponding topic assignment for a given document.  We consider 3 options for the inferential setup: (1) sequential inference, where we first infer a segmentation and then derive a topic assignment given the segmentation (``pipeline''); (2) joint inference, where we jointly optimize for the segmentation and the topic assignment; and (3) independent inference, where we infer the segments regardless of the topics, and the topics regardless of the segments.

In addition, topical segmentation requires a number of segments $k$. This can be decided in a pipeline (i.e., first decide $k$) or jointly (i.e., infer $k$ together with the boundaries). 

Independent and pipeline inference are generally less complex algorithmically, as they allow decomposition of the problem. 

\paragraph{Local Decoding vs. Dynamic Programming.}
Given a desired number of segments, and considering only local scores, we can easily select the optimal segmentation in one linear pass. If we also consider  global scores then we have a structured prediction task that requires dynamic programming in order to be executed in polynomial time, where the degree of the polynomial is decided by the order of dependency.

Given a segmentation, the inference of the optimal topic assignment might require dynamic programming. Since we require adjacent segments to have different topics, greedy local topic inference might not give the optimal topic assignment.

\subsection{Models}

We propose various models and baselines for the task of topical segmentation. Each model is defined as a combination of the possibilities listed above.

\paragraph{Topic-Modeling (TM) Based.} 
\citet{misra2011text} performed segmentation based on topic modeling, where the selected segmentation is that with the highest likelihood, based on a Latent Dirichlet Allocation model (LDA, \citealt{blei2003latent}). 
In this method, we use the likelihood score that the TM gives each segment and find the segmentation that maximizes the product of likelihoods. Inference is equivalent to finding the shortest path in a graph with $n^2$ nodes.

This method jointly infers the number of segments, the segmentation, and topical distributions for each segment. Fixing the number of segments ahead of time requires complex inference as it adds a restriction on the  segmentation.\footnote{\citet{misra2011text} mention that they used a penalty factor for the number of segments, but it remains unclear how it was actually used in the framework, as it introduces dependencies between segment boundaries.}

\paragraph{NSP.}
The approach in the first ChapterCaptor model is to perform linear segmentation based on Next Sentence Prediction (NSP) scores. Using a model that was pretrained for NSP, they further finetune the model with segmented data, where a positive label is given to two subsequent spans in one segment, and a negative label is given to two spans that are in different segments. The spans can be single sentences or some fixed length window. The NSP score is the probability given to the positive label. We denote the NSP score for placing a boundary point after the $n$-th sentence as $NSP_n$.

After computing the NSP score for all $1\leq i \leq n$, the segmentation is derived by selecting the $k$ places where the NSP scores are lowest and placing boundaries there.
We denote this model with {\sc NSP}.

\paragraph{NSP with length penalties.}
The second ChapterCaptor model leverages the assumption that segments tend to have similar lengths. Given data, they compute the expected average length, $L$, and add regularization towards average-length segments. 

Specifically, they use the dynamic formula:

\begin{small}
\begin{multline}
cost(n, k) = \\ \min_{1\leq i\leq n-1} \big( cost(i, k-1) + (1-\alpha) \frac{|n-i-L|}{L} \big) \\+ \alpha \cdot NSP_n
\end{multline}
\end{small}
where $cost(n, k)$ represents the cost of putting a boundary at index $n$ when we already have $k-1$ previous boundaries. $\alpha$ is a hyperparameter controlling the balance between the two factors.

We denote this model with \textit{NSP + L}.

\paragraph{LMPMI.}
Adapting the NSP scores for segmentation seems sub-optimal in domains for which we do not have enough segmented data.
We propose to replace the NSP scores with language-modeling (LM) and Point-wise Mutual Information (PMI) scores. Specifically, for each possible boundary index $i$, we define:

\vspace{-.3cm}
\begin{align}
LMPMI_i = \frac{P_{LM}(x_i, x_{i+1})}{P_{LM}(x_i)\cdot P_{LM}(x_{i+1})}
\end{align}
\vspace{-.3cm}

where the probabilities are the LM probabilities for the sentences together or alone.

These scores can be computed by any pretrained language model, and the log scores replace the NSP scores in both previous methods.  We denote these models with {\sc PMI} and {\sc PMI + L}.




\subsection{Topic Assignment}

\paragraph{Pipeline.}
Given a segmentation for the document and a topic classifier, we can infer a list of topics. We need to find the optimal topic sequence under the constraint of no identical adjacent elements. 

Finding the optimal topic assignment can be formalized as an HMM inference task, which can easily be found using dynamic programming.
Assuming uniform prior probabilities for the topics, the initial state probability is uniform and the transition probabilities are uniform over all states other than the current one. The trained classifier gives us the probabilities of a topic given a segment, $P(t|X)$. With our assumption of uniform topic probabilities, these probabilities are proportional to the emission probabilities. 

\paragraph{Joint Inference.}
As an extension to the previous methods, we propose a formula that takes into account the segment classification scores in addition to the lengths. This is based on the assumption (similar to \citet{misra2011text}), that topically coherent segments will have classification probabilities that are concentrated around the best topic.

Using these scores, we can jointly infer a segmentation and topic assignment. 
We use the following dynamic formula:

\begin{small}
\vspace{-.3cm}
\begin{multline}
cost(n, k, t) = \\ \min_{\substack{1\leq i\leq n-1 \\t' \in T}} \big( cost(i, k-1, t') + \alpha \cdot 
\frac{|n-i-L|}{L} \\ + \beta \cdot \log P(t'|X_i \cdots X_n) \big) + (1- \alpha -\beta) \cdot PMI_n
\end{multline}
\vspace{-.3cm}
\end{small}

where $cost(n, k, t)$ represents the cost of a boundary at index $n$ with $k-1$ previous boundaries and topic $t$ as the last topic. $\alpha, \beta$ are hyperparameters controlling the components.
We denote this model with {\sc PMI + T}.


\subsection{Baseline Models}

As a point of comparison, we also implemented simple baseline models for segmentation and topic selection. These models can be used in a pipeline.  

\paragraph{Uniform Segmentation.}
The simplest way to segment a text is to divide it into equally lengthed segments, given a predetermined number of segments. This method was used by \citet{antoniak2019narrative} and, with slight modifications, by \citet{wu2021recursively}, as it is extremely simple and efficient. 
We set $k$ as the specific document length divided by the average number of tokens per segment in the
development set. 
This baseline is denoted with {\sc Uniform}.

\paragraph{Uniform Topic Selection.}

Given the length of the topic list to extract for the text, we can sequentially sample topics from a uniform distribution over the set of topics. In this case, we can easily avoid repeating topics by giving probability 0 to the previous topic. This too is denoted with {\sc Uniform}.





\section{Experimental Setup}


\subsection{Data}\label{sec:data}

Our data consists of Holocaust survivor testimonies. We received 1000 testimonies from SF. All testimonies were conducted orally with an interviewer, recorded on video, and transcribed as text. The lengths of the testimonies range from $2609$ to $88105$ words, with a mean length of $23536$ words.



\paragraph{Data for the Classifier.} 
The testimonies, originally recorded, were transcribed as time-stamped text. In addition, each testimony recording was divided into segments, typically a segment for each minute. Each segment was indexed with labels, possibly multiple. The labels are all taken from the SF thesaurus.\footnote{\url{https://sfi.usc.edu/content/keyword-thesaurus}} The thesaurus is highly detailed, containing $\sim8000$ unique labels across the segments. 

As some of the labels are very rare, and given the noise in the data, using the full label set directly is dispreferred. 
Instead, we reduced the number of labels
through an iterative process of manual expert annotation and clustering.
The SF thesaurus uses a hierarchical system of labels, ranging from high-level topics (e.g, ``politics'', ``religion and philosophy''), through mid-level labels (e.g., ``camp experiences'', ``ghetto experiences''), to low-level labels (e.g., ``refugee camp injuries'', ``forced march barter''). For the purpose of compiling the list of topics, we focused on mid-level labels. Then, with the help of domain experts from the field of Holocaust studies, we created a list of 29 topics that were deemed sufficiently informative, yet still generalizable across the testimonies. We added the label \emph{NO-TOPIC}, which was used for segments that address technical details of the testimony-giving event (e.g., changing the tape), and do not include  Holocaust-related content.
(the full list can be found in Appendix \ref{appendix: C}).

We filtered out testimonies that were not annotated in the same fashion as the others, for example, testimonies that did not have one-minute segments or ones that skipped segments altogether. We used these testimonies for development and testing. We also filtered out all segments that had more than one label after the label conversion. We ended up with a text classification dataset of $20722$ segments with 29 possible labels.

Since the segments were determined based on time intervals and not content, we cannot use this data as supervision for boundaries, as was done in recent work on segmentation. 

We added to the input texts an extra token to indicate the location within the testimony. We divided each testimony into 10 bins with equal segment counts and added the bin number to the input text. 

\paragraph{Test Data for Segmentation.} 
To compile evaluation and test sets for the topical segmentation problem, we manually segmented and annotated 20 testimonies.  We used testimonies from SF that were not annotated in the same manner as the others, and therefore not used for the classifier.  
The annotation was carried out by two trained annotators, highly proficient in English. 

An initial pilot study to segment testimonies without any prior requirements and no topic list yielded an approximate segment length (the results of these attempts were not included in the training or test data). The approximate length was not used as a strict constraint, but rather as a weak guideline just to align our expectations with the annotators.

The approximate desired average segment length was given to the annotators as well as the final topic list. The first annotator annotated all 20 testimonies, which were used for development and testing. The second annotator annotated 7 documents, used for measuring the inter-annotator agreement. The full annotation guidelines can be found in Appendix \ref{appendix: B}.
 
Altogether, for our test data, we obtained 20 testimonies composed of 1179 segments with topics.  The segment-length ranges from 13 to 8772 words, with a mean length of $\sim 485$. 
We randomly selected 5 testimonies for parameter estimation, and the remaining 15 were used as a test set.

    
    
    \subsection{Classifier Specifics}
    
    The classifier was selected by fine-tuning various Transformer-based models with a classification head. Base models were pretrained by HuggingFace.\footnote{\url{https://pypi.org/project/transformers/}} We experimented with Distilbert, Distilroberta, Electra, Roberta, XLNet, and DeBerta in various sizes.
    For our experiments we chose to use Distilroberta, which showed an accuracy score of $\sim 0.55$, which was close to that of the larger models, doing this with way faster training and inference.
    We trained with a random 80-20 data split on 2 GPUs for $\sim10$ minutes with the \emph{Adam} optimizer for 5 epochs with \emph{batch-size=$16$},  \emph{label-smoothing=$0.01$} and other settings set as default. We selected this classifier for our final segmentation experiments.
    
    We also experimented with a number of linear classifiers. The highest test accuracy we achieved was $\sim 0.46$, which is considerably lower than the one achieved with the neural classifiers. We, therefore, did not use any of the linear models in the final segmentation experiments.

    \subsection{Model Specifics}
    
    From the 20 manually segmented testimonies, we randomly took 5 testimonies a development set for hyperparameter tuning.
    Based on the results on this set, we chose $\alpha=0.8$ for the {\sc PMI + L} model, and $\alpha=\beta=0.2$ for the {\sc PMI + T} model.
    
    
    The LDA topic model was pretrained on the same training data as the classifier's (\S\ref{sec:data}), before running the segmentation algorithm. We trained the LDA model with 15 topics using the Gensim package,\footnote{\url{https://radimrehurek.com/gensim/}} which we also used for the likelihood estimation of text spans given an LDA model. 

    We used HuggingFace's pretrained transformer models for the NSP scores and LM probabilities. We used FNET \citep{lee2021fnet} for NSP and GPT2 \citep{radford2019language} for LM probabilities.
    We experimented with different context sizes $C$ (i.e., how many sentences on each side we use for comparison). We tuned the size parameter on the development set, resulting in $C=3$.
    
    With this setting, the dynamic model with topics takes approximately 50 minutes per testimony, the dynamic model without topics takes approximately 5 minutes per testimony, and the simple gpt2 model takes approximately 2 minutes per testimony, all running with 1 GPU.

\subsection{Evaluation Methods}

Evaluating the classifier component is straightforward since we have labeled data and we can use a held-out test set. We note that the classifier was trained on data that was not divided by topic. We report accuracy scores.
Here we discuss appropriate metrics for the segmentation and topic assignments. 

\paragraph{Segmentation.}
Measuring the quality of text segmentation is tricky. We want to give partial scores to segmentations that are close to the manually annotated ones, so simple Exact-Match evaluation is overly strict. This is heightened in cases like ours, where there is often no clear boundary for the topic changes. For example in one place the witness says \textit{``he helped us later when we planned the escape''}. This sentence comes between getting help (the \textit{Aid} topic) and escaping (the \textit{Escape} topic). We would like to give at least partial scores for boundaries either before or after this sentence. 

Various attempts have been made to alleviate this problem and propose more relaxed measures. Since the notion of ``closeness'' strongly depends on underlying assumptions, it seems hard to pinpoint one specific measure that will perfectly fit our expectations. Following this rationale, we report a few different measures.

The first measure we report is the average F1 score, which counts overlaps in the exact boundary predictions. Another measure we used is average WindowDiff  \citep[WD;][]{pevzner-hearst-2002-critique}, which compares the number of reference boundaries that fall in an interval with the number of boundaries that were produced by the algorithm. We also measured the average Segmentation Similarity \citep[{\sc S-sim};][]{fournier-inkpen-2012-segmentation} and Boundary Similarity  \citep[{\sc B-sim};][]{fournier-2013-evaluating} scores. These scores are based on the number of edits required between a proposed segmentation and the reference, where Boundary Similarity assigns different weights to different types of edits. In F1, {\sc B-sim}, and {\sc B-sim} a higher score is better and in WindowDiff a lower score is better. We used the segeval python package\footnote{\url{https://pypi.org/project/segeval/}} with the default settings to compute all of these measures.
Notably, the window size was set to be the average segment length (in the reference segmentation for the particular testimony) divided by 2.

\vspace{-.1cm}
\paragraph{Topic Assignment.}

One measure we used was python's difflib SequenceMatcher (SM) scores, which are based on the {\it gestalt pattern matching} metric \cite{ratcliff1988pattern}. This metric sums the longest common substrings in a recursive manner, and divides by the total length, attempting to reflect human impression for similarity. In this metric, a higher score means stronger similarity.

Another measure we used is the Damerau–Levenshtein edit distance (Edit, \citealt{10.1145/363958.363994}). This measure defines the distance between two sequences as the minimal number of insertions, deletions, substitutions, or transpositions in order to get from one sequence to the other. Since the number of edits depends on the number of elements in the sequence, we normalized the distance by the number of topics in the reference document.\footnote{We can still get a distance larger than $1$ if the predicted number of topics is larger than the real number. This normalization is commonly known in the literature as {\it word error rate}.} For the Edit distance, lower is better.

\vspace{-.05cm}

\section{Results}

\vspace{-.05cm}
We evaluate our models for both the segmentation and the resulting topic sequence. 

We do not report scores for the LDA-based model since it did not produce a reasonable number of segments, and its runtime was prohibitively long (in previous work, it was run on much shorter text). We also implemented the models with different sizes of GPT2. Observing that the size had no significant effect, we report the results with the base model (\textit{``gpt2''}) only.

We emphasize that our models are only weakly supervised, as the topic classifier was trained with arbitrary boundaries and the topics were implicitly derived from the data.

\vspace{-.1cm}
\paragraph{Annotator Agreement.}
Evaluating on the 7 documents that were annotated by both annotators, we achieve $\textit{Boundary score}=0.324$, $\textit{Sequence Matching}=0.4$ and $\textit{Edit distance}=0.73$. 

In complex structured tasks, the global agreement score is expected to be low. Agreement in these cases is therefore often computed in terms of sub-structures (e.g., attachment score or PARSEVAL F-score in parsing instead of exact match). Since no local scores are common in segmentation tasks, we report only the global scores despite their relative strictness. Compared to the boundary score of uniform-length segmentation (which is much better than random), we can see that the annotator agreement was larger by an order of magnitude. Eyeballing the differences between the annotators also revealed that their annotations are similar. 

We note that the annotators did not always mark the same number of segments (and topics), and this can highly influence the scores. We also note that the annotators worked completely independently and did not adjudicate.

\vspace{-.1cm}
\paragraph{Segmentation.}

\begin{table}
\centering
\begin{tabular}{l|cccc}
\hline \hline
\textbf{Model} & \textbf{F1} & \textbf{WD} & {\sc \textbf{S-sim}} & {\sc \textbf{B-sim}} \\ \hline
{\sc Uniform}  & 0.052 & 0.568 & 0.958 & 0.026 \\ 
NSP + L & 0.04 & 0.584 & 0.958 & 0.02\\
PMI & 0.172 & 0.537 & 0.963 & 0.094\\
PMI + L & {\bf 0.173} & {\bf 0.535} & {\bf 0.964} & {\bf 0.095} \\
PMI + {\sc T} & 0.165 & 0.54 & 0.962 & 0.09 \\
\hline \hline
\end{tabular}
\caption{Segmentation scores. We evaluate PMI-score models with and without length penalties (PMI and PMI + L, respectively). We also evaluate a joint model for segmentation with topics (PMI + {\sc T}), a uniform length segmentor ({\sc Uniform}) and a Next Sentence Prediction segmentor with length penalties (NSP + L). 
For F1, {\sc S-sim} and {\sc B-sim}, higher is better and for WD lower is better. The number of segments is decided using the expected segment length.}
\label{table: 1}
\end{table}

Table \ref{table: 1} presents the results for the segmentation task.
We see that PMI-based models are significantly better than the uniform length segmentation and the NSP-based model. Among the PMI-based models, there is no clear advantage for a specific setting, as the local PMI model is slightly better than the models with global scores. 

Due to the nature of the metrics, specifically how they normalize the values to be between $0$ and $1$, the different measures vary in the significance of the gaps. {\sc S-sim} normalizes by all possible boundaries, so the score will always be high since in most places there is no boundary, even for low quality segmentations. In fact, the cosmetically high values of {\sc S-sim} were one of the incentives for the definition of {\sc B-sim} \citep{fournier-2013-evaluating}. WD uses a sliding window. Therefore, it essentially normalizes by the number of possible boundaries, but, unlike {\sc B-sim}, WD usually counts errors multiple times, resulting in lower scores.

\vspace{-.05cm}
\paragraph{Topic lists.}
\begin{table}[t]
\centering
\begin{tabular}{l|cc}
\hline\hline
\textbf{Model} & \textbf{SM} & \textbf{Edit}  \\ \hline
{\sc Uniform} & 0.138 & 1.13\\ 
{\sc Uniform + Cl} & {\bf 0.378} & {\bf 0.872}\\ 
{\sc NSP + Cl} & 0.369 & 0.875\\
{\sc PMI + Cl} & 0.36 & 0.892\\
{\sc PMI + T} & 0.375 & {\bf 0.872}\\
\hline
{\sc Gold + Cl} & 0.478 & 0.5\\
[0.5ex] 

\hline \hline
\end{tabular}
\caption{Performance of the various models for topic lists. In Sequence Matching (SM) higher is better and for the Edit Distance, lower is better. In all cases, the number of topics was set as the length divided by the expected segment length rounded. The models we evaluate are uniform segmentation, NSP segmentation with length penalties, and PMI segmentation, all with dynamic topic assignment based on the classifier ({\sc Uniform + Cl},  {\sc NSP + Cl} and {\sc PMI + Cl}, respectively), and the joint segmentation and topics model ({\sc PMI + T}). The baseline model is uniform topic generation ({\sc Uniform}), which samples topics independently of the given text, and avoids repeating the previous topic.}
\label{table: 2}
\vspace{-0.35cm}
\end{table}

Table \ref{table: 2} presents our results for the topic assignments produced by our models and the baselines. For comparison, it also presents the scores for topic creation based on the classifier when the real annotated segments are given.

Here we see that the pipeline methods with uniform or NSP segmentation provide slightly better topics than the joint inference model or the simple PMI model. All models based on the classifier perform significantly better than the baselines.


\section{Discussion}

Our results show that topic assignment given the real segmentation {\sc Gold + Cl} gives better topics than all other models. This suggests that a good segmentation does contribute to the topic assignment, which motivates tackling the segmentation and topic assignment jointly, in principle.
The {\sc Gold + Cl} model actually achieves higher topic similarity than the
inter-annotator agreement.
This might be explained by the fact that the {\sc Gold + Cl} model was given the exact number of segments, while this was not specified for the annotators.

Regarding the full models, our results show that the PMI methods show better performance for the segmentation task, compared to previous methods.
This supports our hypothesis that segment boundaries correspond to low mutual information between the segments. This connects to common unsupervised methods and trends, showing that general-purpose self-supervised models can perform strongly on various tasks.

However, we find that the automatic segmentation results do not contribute to the topic assignment (topic assignment scores are comparable with uniform segmentation). It seems then that although the PMI methods show improvement over previous work for the segmentation task, their results are still not sufficient to contribute to the topic extraction.

Within the different PMI models, we see that additional length and topic scores do not yield substantial improvements, neither for the segmentation nor for the topics. This is somewhat surprising and might mean that the sensitivity of our classifier to exact boundaries is low, or that the produced segments did not yet cross a usefulness threshold for topic classification.

Another surprising result is that larger sizes and domain fine-tuning of the GPT2 model do not improve the performance, sometimes actually hurting it. Inspecting the produced segments, it seems that these models do produce meaningful segments with good boundaries, but they don't always match the manual boundaries, as the exact segmentation depends also on the given set of topics. That said, the fact that the models still perform better than baseline models shows that it is possible to produce reasonable segmentations even without specifying a set of topics.

\section{Conclusion}

We presented models for combined segmentation and topic extraction for narratives. We found that: (1) local PMI scores are sufficient to infer a segmentation with better quality than previous models; (2) additional features such as segment lengths and topics seem to have limited influence on the quality of the segmentation; (3) topic lists inferred dynamically given a classifier are not very sensitive to the actual segmentation, allowing the extraction of high-quality topic lists even with uniform segmentation.

Our work addresses the segmentation and topic labeling of text in a naturalistic domain, involving unstructured, transcribed text. Our model can segment noisy texts where textual cues are sparse.  

In addition to the technical contribution of this work, it also makes important first steps in analyzing spoken testimonies in a systematic, yet ethical manner. With the imminent passing of the last remaining Holocaust survivors, it is increasingly important to design methods of browsing and analyzing these testimonies, so as to enable us to use the wealth of materials collected in the archives for studying and remembering the stories.

\section*{Limitations}

Our data for the classification and segmentation are restricted to a specific domain. This limits the generalization of our models to other domains. This is true both regarding the application of the models, as models that use the classifier will require adaptation to a new domain, and regarding the results of the experiments.

Another limitation regards the task of segmentation, as it is not always defined in the same manner and depends on the specific requirements in place. This is true for both supervised and unsupervised methods, since even within a specific domain the optimal segmentation may vary. 

\section*{Ethical Considerations}
We abided by the instructions provided by each of the archives. We note that the witnesses identified themselves by name, and so the testimonies are not anonymous. Still, we do not present in the analysis here any details that may disclose the identity of the witnesses. We intend to release our codebase and scripts, but those will not include any of the data received from the archives; the data and trained models used in this work will not be given to a third party without the consent of the relevant archives.

\section*{Acknowledgments}

The authors acknowledge the USC Shoah Foundation - The Institute for Visual History and Education for its support of this research.
We thank Prof. Gal Elidan, Prof. Todd Presner, Dr. Gabriel Stanovsky, Gal Patel and Itamar Trainin for their valuable insights and Nicole Gruber, Yelena Lizuk, Noam Maeir and Noam Shlomai for research assistance. This research was supported by grants from the Israeli Ministry of Science and Technology and the Council for Higher Education and the Alfred Landecker Foundation.

\bibliographystyle{acl_natbib}
\bibliography{anthology,custom}

\appendix 

\section{Equivalence of PMI and Likelihood} \label{appendix: A}
We have a document $X=x_1, x_2 \cdots, x_n$ which we want to divide into $k$ segments.

We assume that the LM probability for each sentence depends only on the previous sentence and that in the case of a boundary at index $i$, sentence $i$ is independent of all previous sentences. 
Under these assumptions, the segmentation that places boundaries at the places with minimal PMI is the same segmentation that maximized the LM likelihood.

\paragraph{Proof:}

Assume we have a boundary set $B=(i_1, i_2, ... i_k)$.

For any $i \in B$ we have:

\begin{small}
\begin{align*}
PMI(x_i, x_{i-1}) = \frac{P(x_i|x_{i-1})}{P(x_i)} = 1
\end{align*}
\end{small}

Therefore we get:

\begin{small}
\begin{multline}
\argmax_B P(X) = \argmax_B P(X) \cdot \prod_{i=1}^{n} \frac{1}{P(x_i)} \\
=\argmax_B \prod_{i\not\in B} \frac {P(x_{i}|x_{i-1})}{P(x_i)} \prod_{i\in B} \frac{P(x_{i})}{P(x_i)}\\
=\argmax_B \sum _{i\not\in B} logPMI(x_i, x_{i-1}) \\
=\argmax_B \sum _{i=1}^{n} logPMI(x_i, x_{i-1})
\end{multline}
\end{small}

\section{Annotation Guidelines}\label{appendix: B}
\subsection*{Annotation Guidelines for Topical Segmentation}
In this task, we divide Holocaust testimonies into topically coherent segments. The topics for the testimonies were predetermined. We have 29 content topics and a NULL topic. The full list is attached. Each segment has one topic (multi-class, not multi-label), and a change of topic is equivalent to a change of segments.

The segmentation annotation will be as follows:
\begin{itemize}
\item The testimonies are already divided into sentences. A segment change can only be between sentences.
\item Our goal is to annotate segmentations. For this, we will assign a topic for each sentence. Since the main focus is the segment, the topic should be given based on a segment and not a single sentence. 

\item The changing of a tape, if it does not include further information, should not be marked as a separate topic, rather it should be combined with the surrounding topics. If there is a change of topics there then the Overlap should be marked as True over these sentences.
\item Regarding the number of requested segments, we want an approximate average segment length of ~30 sentences. This is a global attribute, as the actual Segment lengths can (and should) vary, depending on the topics. Any single segment should be decided mainly by content and not by constraints regarding the segment lengths.
\item After deciding the segment scope, all sentences can be marked at once. No need to mark them one by one.
\item No sentence should be left without a topic (“None” is also a topic). If the topic is unclear then one should be chosen. It should not be left empty.
\item A “thumb rule” in cases of multiple options is to choose a topic that is more Holocaust-specific. For example, a hiding story about a family member should be assigned to “Hiding” and not to “Family and friendships”.
\end{itemize}

\clearpage
\section{Topic list}\label{appendix: C}

\vspace{.5cm}
\centering
\begin{tabular}{ll}
\multicolumn{1}{c}{\textbf{Topic}} & \multicolumn{1}{c}{\textbf{Description}} \\ \hline
\multicolumn{1}{|l|}{1 Adaptation and survival} & \multicolumn{1}{l|}{\begin{tabular}[c]{@{}l@{}}Any act of finding ways to adapt to the war and persecution \\ and to survive in Ghetto, camps, etc.\end{tabular}} \\ \hline
\multicolumn{1}{|l|}{2 After the war} & \multicolumn{1}{l|}{Not liberation, but post-war life} \\ \hline
\multicolumn{1}{|l|}{3 Aid} & \multicolumn{1}{l|}{Either giving or receiving aid} \\ \hline
\multicolumn{1}{|l|}{4 Antisemitism and persecutions} & \multicolumn{1}{l|}{\begin{tabular}[c]{@{}l@{}}This mostly refers to pre-war episodes, before the ghetto\\ or camps\end{tabular}} \\ \hline
\multicolumn{1}{|l|}{5 Before the war} & \multicolumn{1}{l|}{\begin{tabular}[c]{@{}l@{}}This mostly refers to the opening parts relating the pre-war \\life in the hometown, family, friends, school, etc.\end{tabular}} \\ \hline
\multicolumn{1}{|l|}{6 Betrayals} & \multicolumn{1}{l|}{Any betrayal by friends, neighbors, locals, etc.} \\ \hline
\multicolumn{1}{|l|}{7 Brutality} & \multicolumn{1}{l|}{\begin{tabular}[c]{@{}l@{}}Any acts of brutality, physical or mental during the war - \\intended and performed by someone. To be distinguished from \\hardship which can describe of a certain condition of hardship\end{tabular}} \\ \hline
\multicolumn{1}{|l|}{8 Camp} & \multicolumn{1}{l|}{Any events that take place in the concentration or death camps} \\ \hline
\multicolumn{1}{|l|}{9 Deportations} & \multicolumn{1}{l|}{\begin{tabular}[c]{@{}l@{}}Deportation from the city/village to the ghetto, and from the \\ ghetto to the camps. This includes any forced transport to an \\undesired destination.\end{tabular}} \\ \hline
\multicolumn{1}{|l|}{10 Enemy collaboration} & \multicolumn{1}{l|}{\begin{tabular}[c]{@{}l@{}}Either jews or locals collaborating with the Nazi regime or \\their representatives\end{tabular}} \\ \hline
\multicolumn{1}{|l|}{11 Escape} & \multicolumn{1}{l|}{\begin{tabular}[c]{@{}l@{}}Any escape from hometown, from the ghetto, from prison \\or camps\end{tabular}} \\ \hline
\multicolumn{1}{|l|}{\begin{tabular}[c]{@{}l@{}}12 Extermination/execution/\\death march\end{tabular}} & \multicolumn{1}{l|}{Any event of violent intended killing} \\ \hline
\multicolumn{1}{|l|}{13 Extreme} & \multicolumn{1}{l|}{killing of a child, suicide, surviving a massacre} \\ \hline
\multicolumn{1}{|l|}{14 Family and friendships} & \multicolumn{1}{l|}{Stories involving family members, friends, loved ones} \\ \hline
\multicolumn{1}{|l|}{15 Forced labor} & \multicolumn{1}{l|}{\begin{tabular}[c]{@{}l@{}}Any events taking place in labor camps or as part of forced \\labor\end{tabular}} \\ \hline
\multicolumn{1}{|l|}{16 Ghetto} & \multicolumn{1}{l|}{Any event taking place in the ghetto} \\ \hline
\multicolumn{1}{|l|}{17 Hardship} & \multicolumn{1}{l|}{Any description of physical or mental hardship} \\ \hline
\multicolumn{1}{|l|}{18 Hiding} & \multicolumn{1}{l|}{\begin{tabular}[c]{@{}l@{}}Hiding places, woods, homes while running away or stories of \\being hidden by others (farms, monasteries, etc.),\end{tabular}} \\ \hline
\multicolumn{1}{|l|}{19 Jewish life and faith} & \multicolumn{1}{l|}{\begin{tabular}[c]{@{}l@{}}Any event relating to jewish life and its practices - school, \\prayer, shabbat, synagogue, before, during and after the war\end{tabular}} \\ \hline
\multicolumn{1}{|l|}{20 Liberation} & \multicolumn{1}{l|}{Events relating to allies liberation of camps} \\ \hline
\multicolumn{1}{|l|}{21 Migration} & \multicolumn{1}{l|}{Either pre or post-war migration to other countries} \\ \hline
\multicolumn{1}{|l|}{22 Non Jewish faith} & \multicolumn{1}{l|}{Any mention of non-jewish beliefs, practices etc.} \\ \hline
\multicolumn{1}{|l|}{23 Police/ security /military forces} & \multicolumn{1}{l|}{Events relating to soldiers and police, either enemy or allies} \\ \hline
\multicolumn{1}{|l|}{24 Political activity} & \multicolumn{1}{l|}{Protests, political parties, either for or against Nazis} \\ \hline
\multicolumn{1}{|l|}{25 Prison} & \multicolumn{1}{l|}{Captivity in prison - to be distinguished from camps} \\ \hline
\multicolumn{1}{|l|}{26 Reflection/memory/trauma} & \multicolumn{1}{l|}{} \\ \hline
\multicolumn{1}{|l|}{27 Refugees} & \multicolumn{1}{l|}{\begin{tabular}[c]{@{}l@{}}Mostly the post-war episodes in refugee/displaced persons \\camps\end{tabular}} \\ \hline
\multicolumn{1}{|l|}{28 Resistance and partisans} & \multicolumn{1}{l|}{Any act or resistance, organized or individual} \\ \hline
\multicolumn{1}{|l|}{29 Stills} & \multicolumn{1}{l|}{Presentation of pictures} \\ \hline
\end{tabular}

\end{document}